\documentclass[letterpaper]{article} 
\usepackage{aaai24}  
\usepackage{times}  
\usepackage{helvet}  
\usepackage{courier}  
\usepackage[hyphens]{url}  
\usepackage{graphicx} 
\usepackage{natbib}  
\usepackage{caption} 
\usepackage{algorithm}
\usepackage{algorithmic}
\usepackage{amsfonts}
\usepackage{graphicx}
\usepackage{subfigure}
\usepackage{amsmath}
\usepackage{mathtools}
\usepackage{amssymb}
\usepackage{multicol}

\usepackage{cuted}
\usepackage{newfloat}
\usepackage{listings}
\DeclareCaptionStyle{ruled}{labelfont=normalfont,labelsep=colon,strut=off} 
\floatstyle{ruled}
\newfloat{listing}{tb}{lst}{}
\floatname{listing}{Listing}
\title{Reward Machine Inference for Robotic Manipulation}
\author{
    Mattijs Baert, Sam Leroux, Pieter Simoens
}
\affiliations{
IDLab, Department of Information Technology at Ghent University - imec \\
Technologiepark 126 \\ 
B-9052 Ghent, Belgium \\
\{mattijs.baert, sam.leroux, pieter.simoens\}@ugent.be
}

\begin{document}

\maketitle

\begin{abstract}
Learning from Demonstrations (LfD) and Reinforcement Learning (RL) have enabled robot agents to accomplish complex tasks. Reward Machines (RMs) enhance RL's capability to train policies over extended time horizons by structuring high-level task information. In this work, we introduce a novel LfD approach for learning RMs directly from visual demonstrations of robotic manipulation tasks. Unlike previous methods, our approach requires no predefined propositions or prior knowledge of the underlying sparse reward signals. Instead, it jointly learns the RM structure and identifies key high-level events that drive transitions between RM states. We validate our method on vision-based manipulation tasks, showing that the inferred RM accurately captures task structure and enables an RL agent to effectively learn an optimal policy.

\end{abstract}

\section{Introduction}
Combining Learning from Demonstrations (LfD) with Reinforcement Learning (RL) has empowered artificial agents to learn complex tasks from human examples in areas such as video games \cite{hester2018deep}, board games \cite{silver2016mastering}, and robotics \cite{argall2009survey, abbeel2010autonomous}. LfD enhances RL by improving sample efficiency, which accelerates learning, and by reducing the effort required from human developers to manually specify precise task objectives. However, many LfD approaches struggle with long-duration tasks as they focus on directly learning a control policy from demonstrations \cite{DBLP:conf/icra/ZhangMJLCGA18,DBLP:conf/rss/MandlekarXMS020} or inferring a reward function \cite{ho2016generative, abbeel2004apprenticeship, baert2023maximum}. Integrating abstract task structures, by decomposing complex tasks into manageable sub-tasks and providing structured guidance to the agent, has enabled RL to address long-horizon tasks more effectively \cite{baert2024learningtask, camacho2021reward}.
\\ \\
Reward machines (RMs) \cite{icarte2022reward} provide a framework for defining such abstract task structures by encoding high-level task objectives in a structured, automata-inspired format. Originally RMs were developed to extend RL for environments with non-Markovian rewards, by enhancing the agent’s state space with an abstract state layer that allow agents to retain memory of past actions. However, recent research \cite{camacho2021reward} has demonstrated that RMs can also enhance performance in fully Markovian tasks by converting sparse task rewards into a denser reward signal, thereby offering more consistent guidance as the policy advances through abstract states. Despite their advantages, most current approaches require RMs to be manually specified by domain experts.
\\ \\
In this work, we address this gap by focusing on learning reward machines directly from visual demonstrations in robotic manipulation tasks. The RM framework depends on a set of propositions that encode high-level properties of the environment and a labeling function $L$ that assigns truth values to these propositions based on the current environment state. Unlike prior approaches, which assume these propositions or feature detectors are predefined \cite{toro2019learning,xu2021active,verginis2024joint, camacho2021reward}, our approach jointly learns the RM structure and the mapping from environment states to Boolean propositions. Our method begins by capturing visual demonstrations. We leverage the observation that sub-goals are visited more frequently in expert demonstrations compared to other states \cite{ghazanfari2020sequential,baert2024learning}.
To identify these sub-goals, each frame is mapped to a low-dimensional feature vector. Through clustering, similar states are grouped and prototypical states (representing a certain sub-goal) can be identified. Finally, an RM is constructed capturing the sequential task structure from the demonstrations. This approach advances LfD by enabling the automatic inference of structured task representations, which can effectively guide RL agents in long-horizon tasks. We evaluate our method on a set of vision-based robotic manipulation tasks, demonstrating its effectiveness.

\section{Related Work}
\label{sec:related}
There has been extensive research on integrating formal methods into RL. For example, several works adopt temporal logics to specify complex, long-horizon tasks \cite{li2017reinforcement,kuo2020encoding,xiong2022constrained,voloshin2022policy}, while others investigate the use of automata, such as RMs, to formalize task specifications \cite{araki2019learning,araki2021logical,icarte2022reward}. A common assumption in these approaches is that high-level knowledge, in the form of automata or temporal logic specifications, is available a priori. However, in real-world scenarios, this knowledge is often implicit and must be inferred from data.
\\
\\ 
Regarding the inference of temporal logics from data, many methods focus on learning temporal logic formulas using both positive and negative examples \cite{kong2016temporal, bombara2021offline}. However, in the context of LfD, only positive examples (i.e., demonstrations) are typically available. Some works address learning linear temporal logic (LTL) from only positive examples \cite{shah2018bayesian, roy2023learning}, but these approaches usually assume that for each time step in the provided trajectories, the truth valuation of a set of Boolean propositions is known. More recent efforts aim to learn both a mapping from states to atomic propositions and the formula structure \cite{baert2024learning}, although applying this in continuous state spaces remains challenging.
\\ 
\\ 
In the domain of learning reward machines, most approaches focus on jointly learning both the RM and the policy through environment interaction. For example, discrete optimization can be used to learn an RM that decomposes the task into subproblems, such that combining their optimal memoryless policies yields an optimal solution for the original task \cite{toro2019learning}. \citet{xu2020joint} propose an iterative method that alternates between automaton inference, used to hypothesize RMs, and RL to optimize policies based on the current RM candidate. Inconsistencies between the hypothesis RM and observed trajectories then trigger re-learning. \citet{verginis2024joint} extend this approach to settings with partially known semantics. Additionally, \citet{xu2021active} and \citet{dohmen2022inferring} leverage active automata inference algorithms, such as the L* algorithm, to learn RMs. Although these approaches do not rely on expert demonstrations, they assume access to the reward function of the underlying MDP as well as the labeling function that maps states to propositions.
Closest to our work is the method proposed by \citet{camacho2021reward}, which learns RMs for vision-based robotic manipulation tasks. However, this approach also assumes access to a predefined mapping between low-level states and high-level propositions, limiting its applicability in more general settings where such a mapping is not readily available.

\section{Background}
\subsection{Reinforcement Learning}
A Markov decision process (MDP) \cite{bellman1957markovian} models sequential decision-making with the following components: a state space $\mathcal{S}$, an action space $\mathcal{A}$, a discount factor $\gamma \in [0,1]$, a transition distribution $p(s^{\prime} \mid s, a)$ describing the probability of reaching state $s^\prime$ from state $s$ when taking action $a$, an initial state distribution $\mathcal{I}(s)$, and a reward function $R: \mathcal{S} \times \mathcal{A} \mapsto \mathbb{R}$, which defines a scalar reward for each state-action pair. An agent interacts with the environment at discrete timesteps $t$, generating trajectories $\tau = (s_0, ..., s_{T-1})$ of length $T$. The RL objective is to find an optimal policy $\pi$ that maximizes the expected sum of discounted rewards: $\max_{\pi}\mathbb{E}_{\pi} \sum_{t=0}^{T-1}\gamma^t R(s_t,a_t)$. 
\\
\\
In this work, we use off-policy deep Q-learning (DQN) \cite{mnih2015human} to train a policy $\pi$ that selects actions by maximizing a Q-function: $\arg\max_{a \in \mathcal{A}}Q_{\theta}(s,a)$. Here, $Q_{\theta}(s,a)$, a neural network with parameters $\theta$, estimates the expected cumulative reward for taking action $a$ in state $s$. DQN updates $\theta$ by minimizing the temporal-difference (TD) error between the current Q-value estimate and a target value $y_t$. For each training step, transitions $(s_t, a_t, r_t, s_{t+1})$ are sampled from a replay buffer and the loss is defined as:
\begin{equation}
    \mathcal{L}(\theta) = \textrm{Huber}(Q_{\theta}(s_t,a_t) - y_t) 
\end{equation}
where
\begin{equation}
    y_t = r(s_t, a_t, s_{t+1}) + \gamma \max_{a^{\prime}} Q_{\theta}(s_{t+1}, a^{\prime}).
\end{equation}
Here, $a^{\prime}$ represents the set of all available actions.

\subsection{Reward Machines}
A Reward Machine (RM) \cite{icarte2022reward} is defined as a tuple $\mathcal{R}_{\textrm{\textit{AP}}, \mathcal{S},\mathcal{A}} = \langle U, u_0, \delta_u, \delta_r \rangle$, given a set of atomic propositions \textit{AP}, a set of environment states $\mathcal{S}$, and a set of actions $\mathcal{A}$. Each proposition  $p \in \textrm{\textit{AP}}$ has a truth value of either true or false and represents a specific piece of information about the environment, such as object properties, agent statuses, or environmental conditions. The negation of a proposition $p$ is denoted as $\neg p$. Propositions can be combined into more complex logical expressions using conjunctions ($\wedge$) and disjunctions ($\vee$). $U$ is a finite set of states, with $u_0 \in U$ representing the initial state.
The state-transition function $\delta_u$ maps pairs $(u, \textrm{\textit{AP}})$ to new states in $U$, while the reward-transition function $\delta_r$ maps state transitions $(u, u^{\prime})$ to real-valued rewards in $\mathbb{R}$. At each time step $t$, the RM receives a truth assignment $\mathbf{p}_t$, which includes the propositions in \textit{AP} that are true in the current environment state $s_t$. We can replace the standard reward in an MDP by an RM, creating a MDPRM. This requires a labeling function $L: \mathcal{S} \rightarrow 2^{\mid\textrm{\textit{AP}}\mid}$ that assigns truth values to the propositions in \textit{AP} based on the environment state. The state of the RM is updated every time step of the environment. If the RM is in state $u$ and the agent takes action $a$ to transition from environment state $s$ to $s^{\prime}$, the RM transitions to state $u^{\prime} = \delta_u(u,L(s^{\prime}))$ and the agent receives a reward $r = \delta_r(u,u^{\prime})$. A policy $\pi(s,u)$ for an MDPRM is conditioned both on the environment state and the RM state. This setup enables the modeling of non-Markovian reward functions within an MDPRM, as different histories of environment states can be distinguished by elements of a finite set of regular expressions over \textit{AP}. Consequently, RMs can yield different rewards for identical environment transitions $(s,a,s^{\prime})$, depending on the agent's prior state history.

\section{Reward Machine Inference}
\begin{figure*}
    \includegraphics[width=\linewidth]{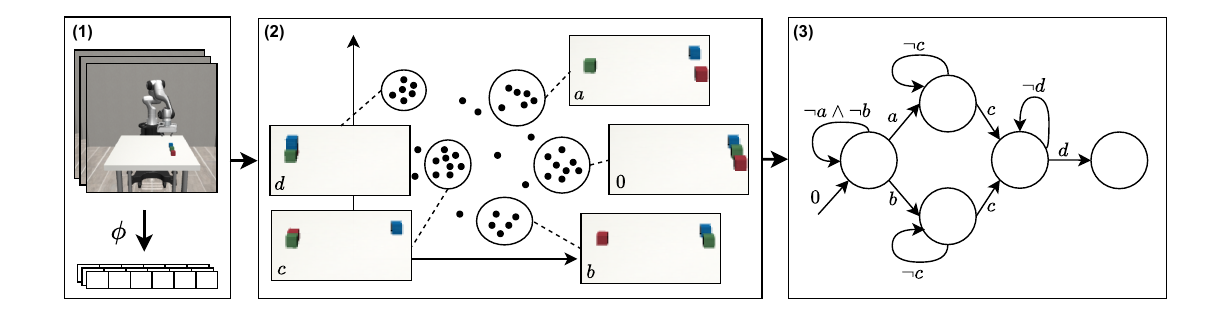}
    \caption{Overview of the proposed method applied to the task of building a predefined pyramid. (1) Visual demonstrations are captured, and feature embeddings are extracted using a pre-trained model $\phi$. (2) Sub-goals are inferred by clustering the feature vectors obtained from the demonstrations. (3) An RM is constructed, capturing the valid temporal ordering and transitions between the inferred sub-goals. The $0$-prototype corresponds with the initial RM state.}
    \label{fig:rm-inference-overview}
\end{figure*}
In this section, we describe the process of inferring an RM from a set of high-dimensional demonstrations. The method consists of four steps: (1) capturing demonstrations, (2) extracting feature representations, (3) inferring sub-goals through clustering, and (4) constructing the reward machine. An overview of this process, applied to the task of building a predefined pyramid, is depicted in Fig. \ref{fig:rm-inference-overview}.

\subsection{Capturing Demonstrations}
The first step is to capture a set of demonstrations from an expert performing the task. A camera with full observability of the workspace records these demonstrations, which results in a set of captured trajectories, denoted as $\mathcal{D} = \{\tau_0, \tau_1, \dots\}$, where each trajectory $\tau_i$ represents a sequence of observed states over time. Since an image frame contains all the necessary information for the agent to make decisions, we define a state $s_t \in \mathcal{S}$ as the image frame at time step $t$. Each state is represented as a tensor in $\mathbb{R}^{256 \times 256 \times 3}$, where the dimensions refer to the height, width, and RGB color channels of the image. Additionally, we capture the same demonstrations using a top-down camera, however, these are only utilized during the policy training phase (as described in the next section).

\subsection{Extracting Feature Representations}
To process the demonstrations, each frame $s_t$ is cropped to retain only the most relevant portion of the workspace, such as the tabletop area where task-relevant objects are located (see Fig. \ref{fig:rm-inference-overview}). After cropping, we extract a feature representation from each frame. This process transforms the image frames into a lower-dimensional feature space that retains the essential visual information. The extracted feature representation of a state $s_t$ is denoted by $\phi(s_t)$.

\subsection{Inferring Sub-Goals through Clustering}
To identify sub-goals, we leverage the insight that true sub-goals occur more frequently in expert demonstrations than other states \cite{ghazanfari2020sequential,baert2024learning}. Consequently, clustering can be used to detect high-density regions in the state space. Sub-goals are inferred by clustering the feature vectors obtained from the previous step, using the DBSCAN algorithm \cite{ester1996density}. DBSCAN is robust to noise and does not require prior knowledge of the number of clusters, which is crucial as the number of sub-goals in the task is unknown. Let $\phi(\mathcal{D}) = {\phi(s) \mid s \in \tau, \tau \in \mathcal{D}}$ represent the set of all feature vectors extracted from all demonstrations. DBSCAN is applied to $\phi(\mathcal{D})$, yielding a set of $k$ clusters $\{C_0, C_1, \dots, C_{k-1}\}$. The cluster center for each cluster $C_i$ is defined as $\mu_i$:
\begin{equation}
    \mu_i = \dfrac{1}{\mid C_i \mid} \sum_{\phi(s) \in C_i} \phi(s).
\end{equation}
For each cluster $C_i$, we identify a prototypical state $s_i^{*}$, which is the state whose feature vector is closest to the cluster center:
\begin{equation}
    s_i^{*} = \arg\min_{s \in C_i}\mid\mid \phi(s) - \mu_i\mid\mid_2.
\end{equation}
Here, $||\cdot||_2$ represents the Euclidean norm. Although clustering is performed in feature space, each prototypical state corresponds to an interpretable image, providing a human-understandable representation of each sub-goal.
\begin{algorithm}[tb]
\caption{Inferring the state transition function $\delta_u$ from the set of abstract demonstrations $\hat{\mathcal{D}}$}
\label{alg:delta_u_inf}
\begin{algorithmic}
\REQUIRE set of abstract demonstrations $\hat{\mathcal{D}}$
\STATE $\delta_u(u, \cdot) \leftarrow \textrm{undefined} \;\;\;\; \forall u \in U$
\FOR{$\hat{d} \in \hat{\mathcal{D}}$}
\STATE $u \leftarrow u_0$
\FOR{$\mathbf{p} \in \hat{d}$}
\FOR{$p_i \in \mathbf{p}$}
\IF{$\delta_u(u,p_i) = \textrm{undefined}$}
\STATE $\delta_u(\hat{u}, \hat{p}) \leftarrow \begin{cases}
      u_i & \text{if} \; \hat{u} = u \text{ and } \hat{p} = p_i \\
      \delta_u(\hat{u}, \hat{p}) & \text{otherwise}
    \end{cases}$
\ENDIF
\STATE $u \leftarrow \delta_u(u, p_i)$
\ENDFOR
\ENDFOR
\ENDFOR
\RETURN $\delta_u$
\end{algorithmic}
\end{algorithm}
\ \\

\subsection{Constructing the Reward Machine}
The set of prototypical states forms the basis for defining the reward machine states. Let $U = \{u_0, u_1, \dots, u_{k-1}\}$ represent the set of RM states, where each state $u_i$ corresponds to a prototypical state $s_i^*$. We build the set of atomic propositions \textit{AP} by defining a proposition $p_i$ for each prototypical state $s_i^*$. The truth evaluation of each proposition $p_i \in \textrm{\textit{AP}}$ is determined based on the Euclidean distance between the feature representation of the current state $\phi(s)$ and the feature representation of the corresponding prototypical state $\phi(s_i^*)$. Formally, for a given proposition $p_i \in \textrm{\textit{AP}}$, corresponding to the sub-goal represented by prototypical state $s_i^*$ and RM state $u_i$, the proposition is true if the Euclidean distance between $\phi(s)$ and $\phi(s_i^*)$ is less than a predefined threshold $\kappa$. The labeling function can then be expressed as:
\begin{equation}
    L(s) = \{p_i \mid ||\phi(s) - \phi(s_i^*)||_2 < \kappa, \forall p_i \in \textrm{\textit{AP}} \}.
    \label{eq:labeling}
\end{equation}
An abstract demonstration $\hat{d}$ associated to a demonstration $d$ can be defined as a sequence of sets of propositions $\mathbf{p}_0, \mathbf{p}_1,...,\mathbf{p}_{n-1}$, where $\mathbf{p}_t$ is the abstraction of state $s_t$ into \textit{AP} by $L$, for each $t \in \{0,\dots,T-1\}$. The set of abstract demonstrations $\hat{\mathcal{D}}$, thus, defines the directed connections between RM states. Transitions between abstract states observed in the demonstration should be reflected into the state-transition function $\delta_u$. The inference of the state transition function $\delta_u$ is formalized in Algorithm \ref{alg:delta_u_inf}. Given the definition of the labeling function (Eq. \ref{eq:labeling}), multiple propositions may hold true for a single state. However, this would imply that the RM occupies multiple states simultaneously, which is not feasible. To avoid this, the hyperparameter $\kappa$ should be tuned so that, for each state $s_t$, at most one proposition is true.
\\
\\
Next, we need to define the reward transition function $\delta_r$ to guide the agent toward reaching the goal states. A naive approach would be to assign a reward of 1 when the agent reaches a goal state in the RM and 0 otherwise. However, this creates a highly sparse reward signal, making it difficult for the agent to learn efficiently. To address this, we use potential-based reward shaping \cite{ng1999policy}, which helps to construct a denser reward function while maintaining the same optimal behavior as the original sparse reward. The idea is to provide intermediate rewards that encourage progress toward the goal, making training more efficient. Inspired by \cite{camacho2021reward}, we define the reward function as follows:
\begin{align}
    \Psi(u) &= \gamma^{d_{\textrm{goal}}(u)} \\
    \delta_r(u, u^{\prime}) &= \delta_{r}^{\prime}(u, u^{\prime}) + \gamma \Psi(u^{\prime}) - \Psi(u),
    \label{eq:rm_reward}
\end{align}
where $\Psi$ is the potential function, $d_{\textrm{goal}}$ represents the shortest distance from the current RM state $u$ to the goal state in terms of edges and $\delta_{r}^{\prime}$ is the original sparse reward function. This shaped reward has the property that the reward is negative when the agent moves away from the goal in the RM graph, is slightly less negative when the agent stays in the same state, is zero when the agent moves closer to the goal and evaluates to 1 when the agent reaches the goal state. 

\section{DQN on the Inferred Reward Machine}
\begin{figure}
    \includegraphics[width=\linewidth]{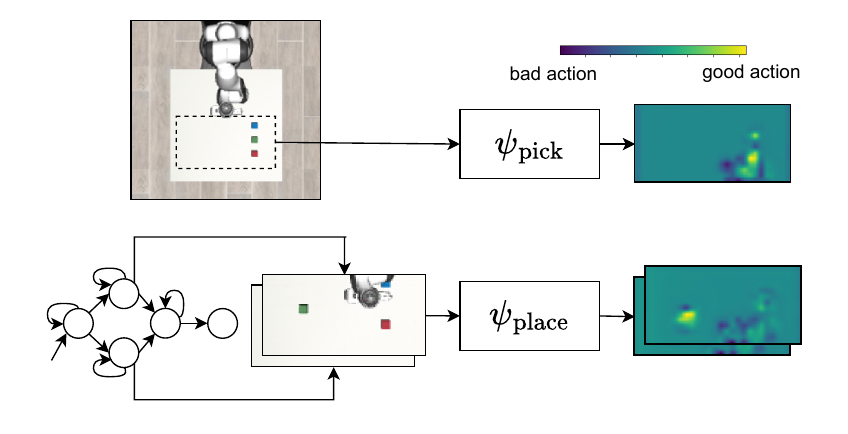}
    \caption{Overview of our DQN formulation. Each action consists of a picking operation at pixel coordinates $(u_{\textrm{pick}},v_{\textrm{pick}})$ followed by a placing operation at $(u_{\textrm{place}}, v_{\textrm{place}})$. The Q-function $Q_{\theta}(s,a)$ is modeled using two FCN's: $\psi_{\textrm{pick}}$ and $\psi_{\textrm{place}}$ which generate pixel-wise Q-value maps for their respective actions.}
    \label{fig:dqn}
\end{figure}
The action space is defined by a pick-and-place primitive, parameterized by four pixel coordinates: two for the pick location $(u_{\textrm{pick}},v_{\textrm{pick}})$ and two for the place location $(u_{\textrm{place}}, v_{\textrm{place}})$. Through camera calibration and depth measurements, each pixel coordinate is mapped accurately to the robot's coordinate system, ensuring precise execution of these actions.
For RM inference, images are captured from a front-view camera, while control actions use a top-down view of the workspace. The top-down perspective aligns directly with the workspace plane, reducing distortion and occlusions and facilitating accurate pixel-to-coordinate mappings. In contrast, front-view images are used for inferring sub-goals, as they avoid occlusions caused by the robot and provide a fuller view of the workspace.
\\
\\
Inspired by prior work in robotic manipulation \cite{camacho2021reward, zeng2022robotic}, we define our deep Q-function $Q_{\theta}(s,a)$  using two fully convolutional networks (FCNs) \cite{long2015fully}, $\psi_{\textrm{pick}}$ and $\psi_{\textrm{place}}$. Each FCN receives a top-down image and outputs a dense, pixel-wise map of Q-values for all pick-and-place actions. To stabilize training, we only consider the Q-values inside a rectangular region corresponding to the table area, as picking or placing outside this region is irrelevant. Both FCNs share the same architecture based on ResNet50 \cite{he2015deep} pre-trained on ImageNet \cite{deng2009imagenet}. We use an intermediate feature map from the output of the second residual block, then pass it through two convolutional layers that reduce the channels to one. Finally, the output is bilinearly upsampled to match the input resolution. During training, we compute gradients only for the Q-value of the executed action’s pixel location, backpropagating zero loss for all other pixels. All parameters, including those in the pre-trained ResNet layers, are updated.
\\ \\
Following the DQRM approach \cite{icarte2022reward}, we train a discrete Q-function per RM state, each with a separate experience buffer for storing state-specific interactions and demonstrations. At the end of each episode, we update all Q-functions by sampling batches from their corresponding buffers. The input to the pick-action network $\psi_{\textrm{pick}}$ is the current observation, while for $\psi_{\textrm{place}}$, we pass a top-down view prototype for each valid transition from the current RM state. This allows the network to focus on areas where blocks are expected in the next state, aiding in goal-oriented placement. Fig. \ref{fig:dqn} depicts an overview of our DQN formulation applied to the pyramid task introduced earlier. Fig. \ref{fig:dqn} is captured in the beginning of an episode with each block in its initial position and the RM in its initial state. There are two valid transitions from the initial RM state, resulting in the place network $\psi_{\textrm{place}}$ receiving two corresponding top-down prototypes. The Q-value maps in Fig. \ref{fig:dqn} reveal that the pick network assigns high Q-values to the locations of the red and green blocks, indicating these are favorable pick locations. Similarly, the place network predicts high Q-values at the designated goal locations for each block.

\section{Results}
\begin{figure}[t]
    \centering
    \includegraphics[width=\linewidth]{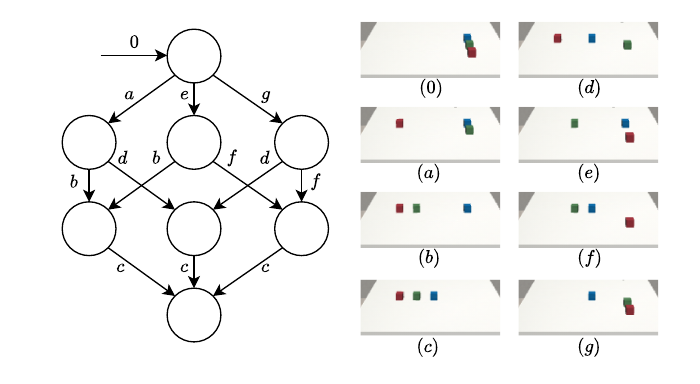}
    \caption{Inferred RM for the Place-3 task (left) and the corresponding state prototypes (right). We omitted the self-loop transitions for clarity.}
    \label{fig:3b_2}
\end{figure}
\begin{figure}[t]
    \centering
    \includegraphics[width=\linewidth]{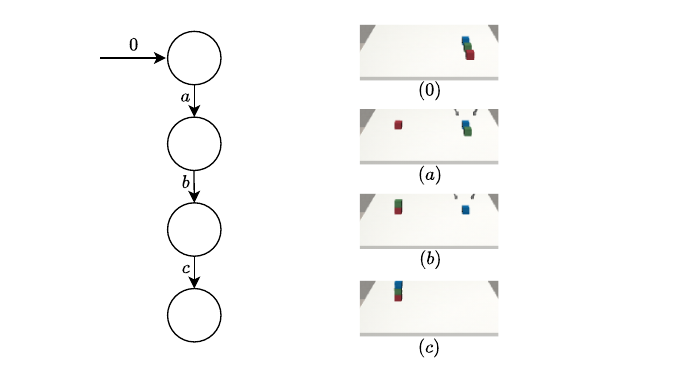}
    \caption{Inferred RM for the Stack-3 task (left) and the corresponding state prototypes (right). We omitted the self-loop transitions for clarity.}
    \label{fig:3b_3}
\end{figure}

We evaluated our method on object manipulation tasks using unstructured human demonstrations within a simulated environment, employing a Franka Emika Panda robot in the robosuite simulation framework \cite{robosuite2020}. Five block-based manipulation scenarios were designed: Stack-2, where two blocks are stacked in a predefined order; Place-2, placing two blocks at specific locations; Pyramid-3, building a predefined three-block pyramid (Fig. \ref{fig:rm-inference-overview}); Stack-3 (Fig. \ref{fig:3b_3}) and Place-3 (Fig. \ref{fig:3b_2}), similar to Stack-2 and Place-2, but involving three blocks. The number of demonstrations collected varied with task complexity. For example, Stack-3 required only a single demonstration due to the unique path from the initial to terminal RM state. In contrast, Place-3 required six demonstrations to cover all possible variations in sub-goal sequences. Demonstrations were generated by controlling the robot through an algorithmically defined expert policy, ensuring high-quality demonstrations that fully represent task variations. For RM inference, the feature extractor $\phi$ is parameterized by a ResNet-50 model \cite{he2015deep} pre-trained on ImageNet \cite{deng2009imagenet}. DQN on the inferred RM was conducted with a fixed batch size of 16, using the Adam optimizer \cite{kingma2014adam} and a learning rate of 0.0001. An $\epsilon$-greedy exploration strategy is used, with $\epsilon$ initialized at $0.7$ and decaying exponentially to $0.1$ over training.
DBSCAN requires two parameters: $\epsilon_{\textrm{cluster}}$, which defines the maximum distance between two points for one to be considered in the neighborhood of the other, and \textit{min points}, the minimum number of points needed to establish a dense region. For each scenario, we tuned these parameters along with the threshold $\kappa$ for matching states with prototypes. Table \ref{tab:parameters} provides an overview of the tuned parameters. 
\setlength{\tabcolsep}{5pt}
\renewcommand{\arraystretch}{1.1}
\begin{table}[]
    \centering
    \begin{tabular}{c|cccc}
         & $\kappa$ & $\epsilon_{\textrm{cluster}}$ & \textit{min points} & \# demonstrations \\
         \hline
         Stack-2 & 3.0 & 0.8 & 15 & 1 \\ 
         Place-2 & 1.9 & 1.0 & 20 & 2 \\ 
         Pyramid-3 & 2.3 & 1.2 & 30 & 2\\ 
         Stack-3 & 3.0 & 1.8 & 40 & 1\\ 
         Place-3 & 1.9 & 1.7 & 110 & 6\\ 
    \end{tabular}
    \caption{Tuned parameters and the number of demonstrations used for the different scenarios.}
    \label{tab:parameters}
\end{table}
\begin{figure*}
\begin{center}
    \subfigure{\includegraphics[width=0.19\textwidth]{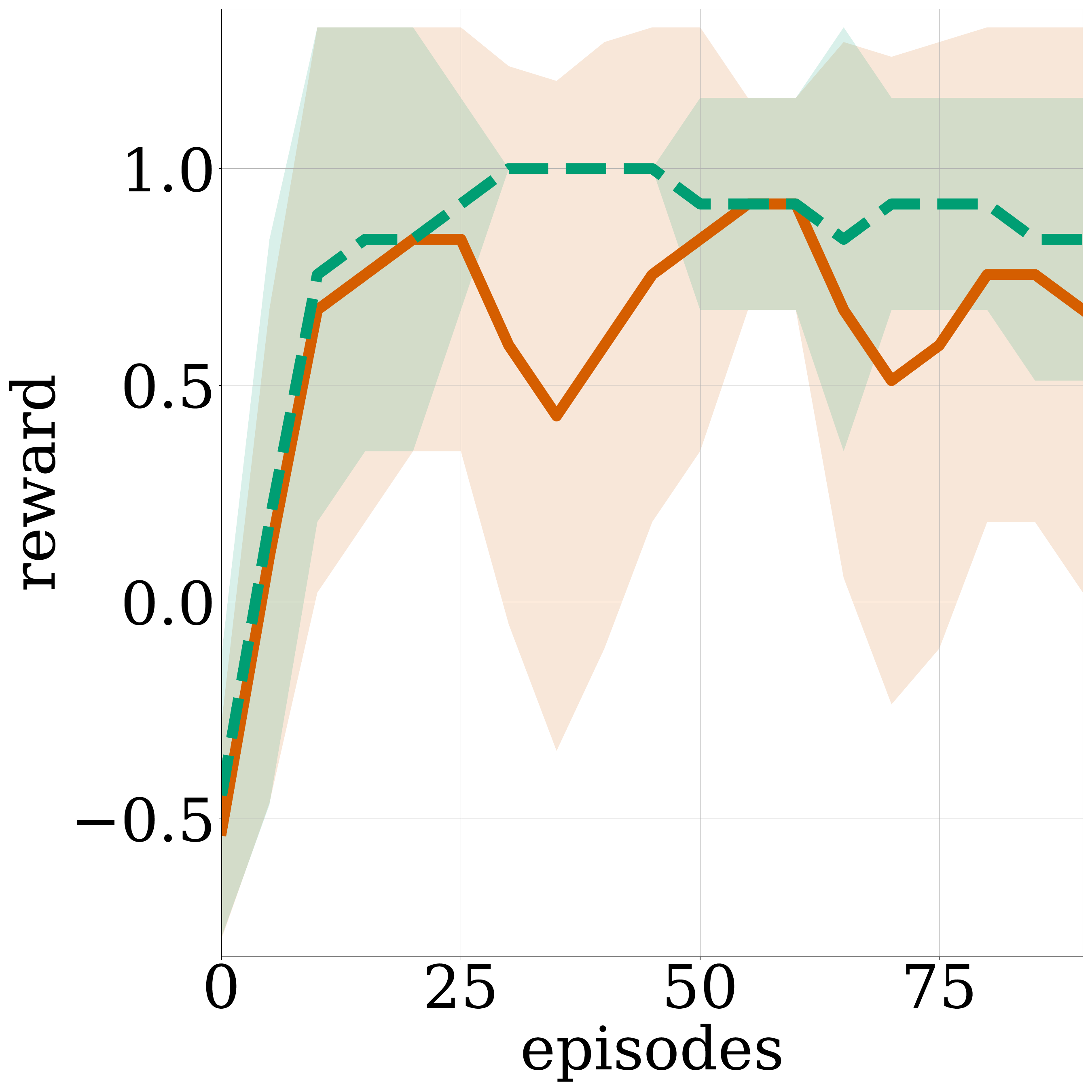}}
    \subfigure{\includegraphics[width=0.19\textwidth]{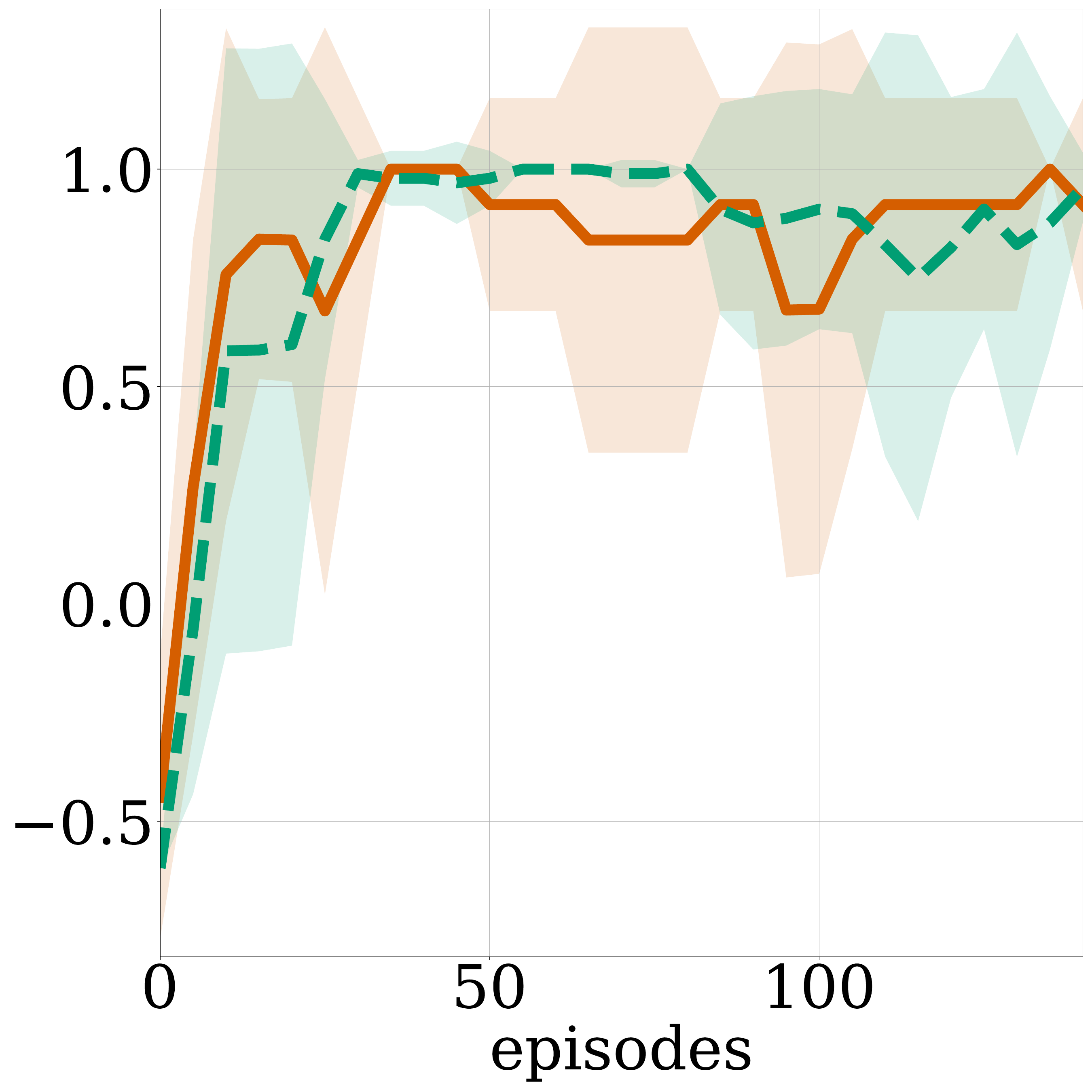}}
    \subfigure{\includegraphics[width=0.19\textwidth]{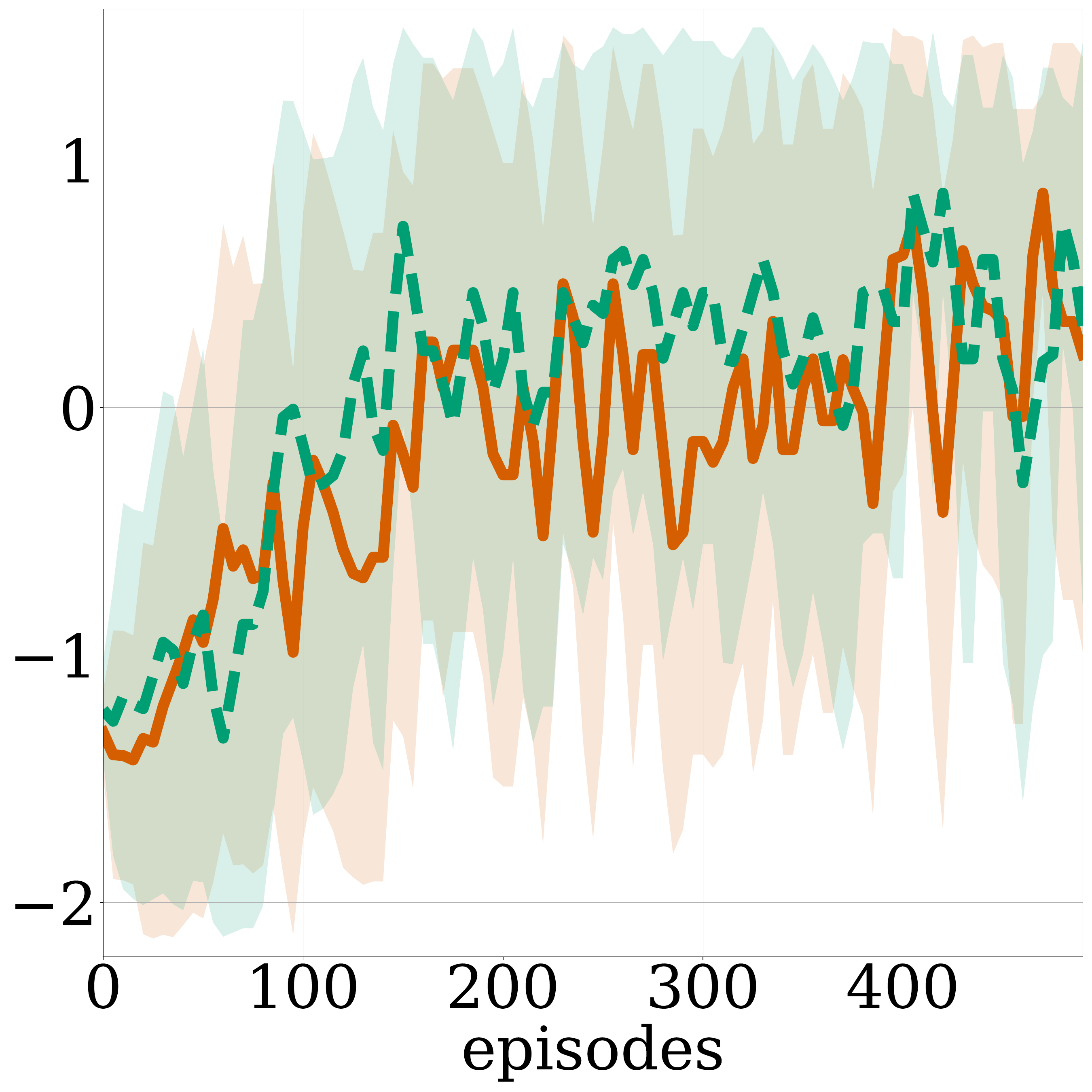}}
    \subfigure{\includegraphics[width=0.19\textwidth]{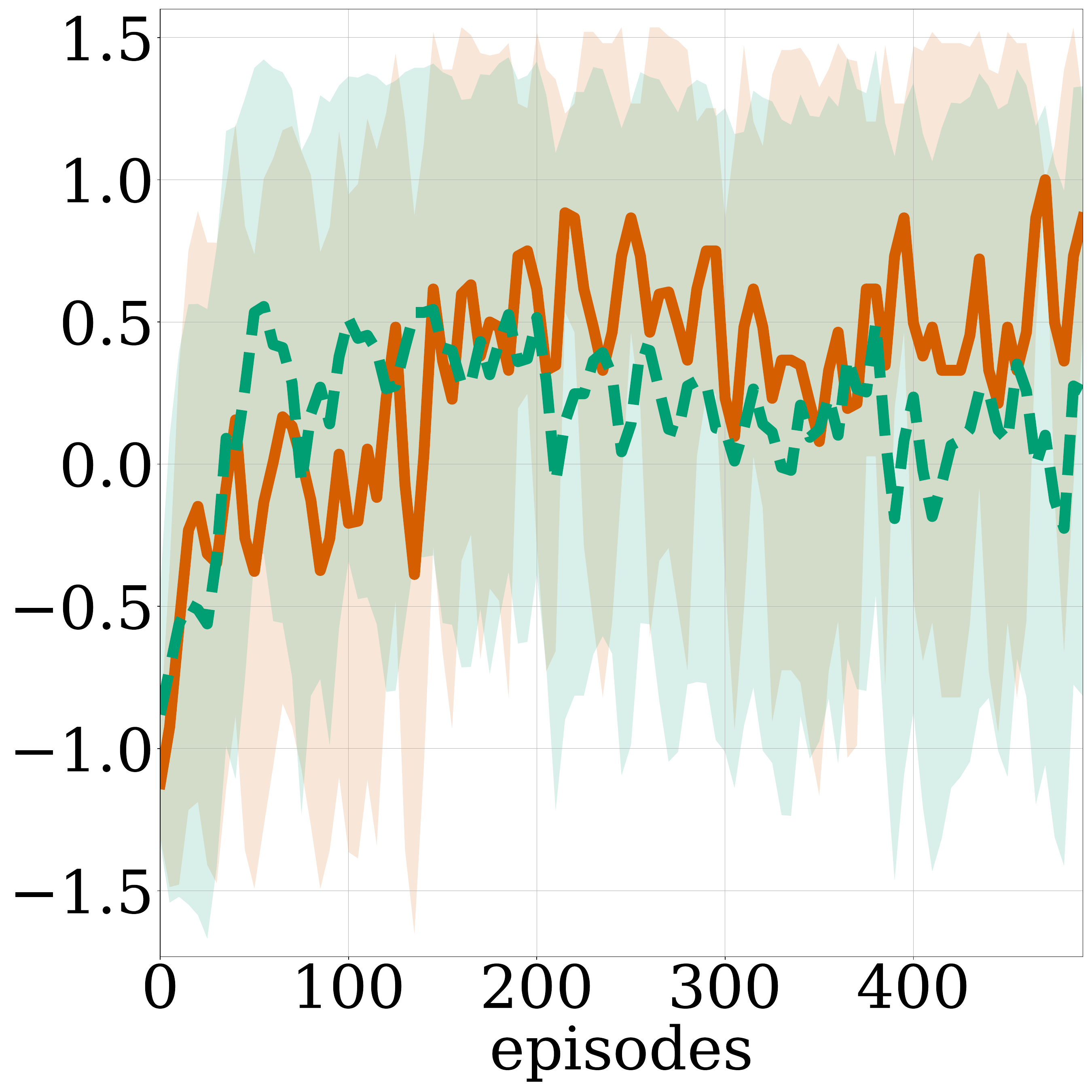}}
    \subfigure{\includegraphics[width=0.19\textwidth]{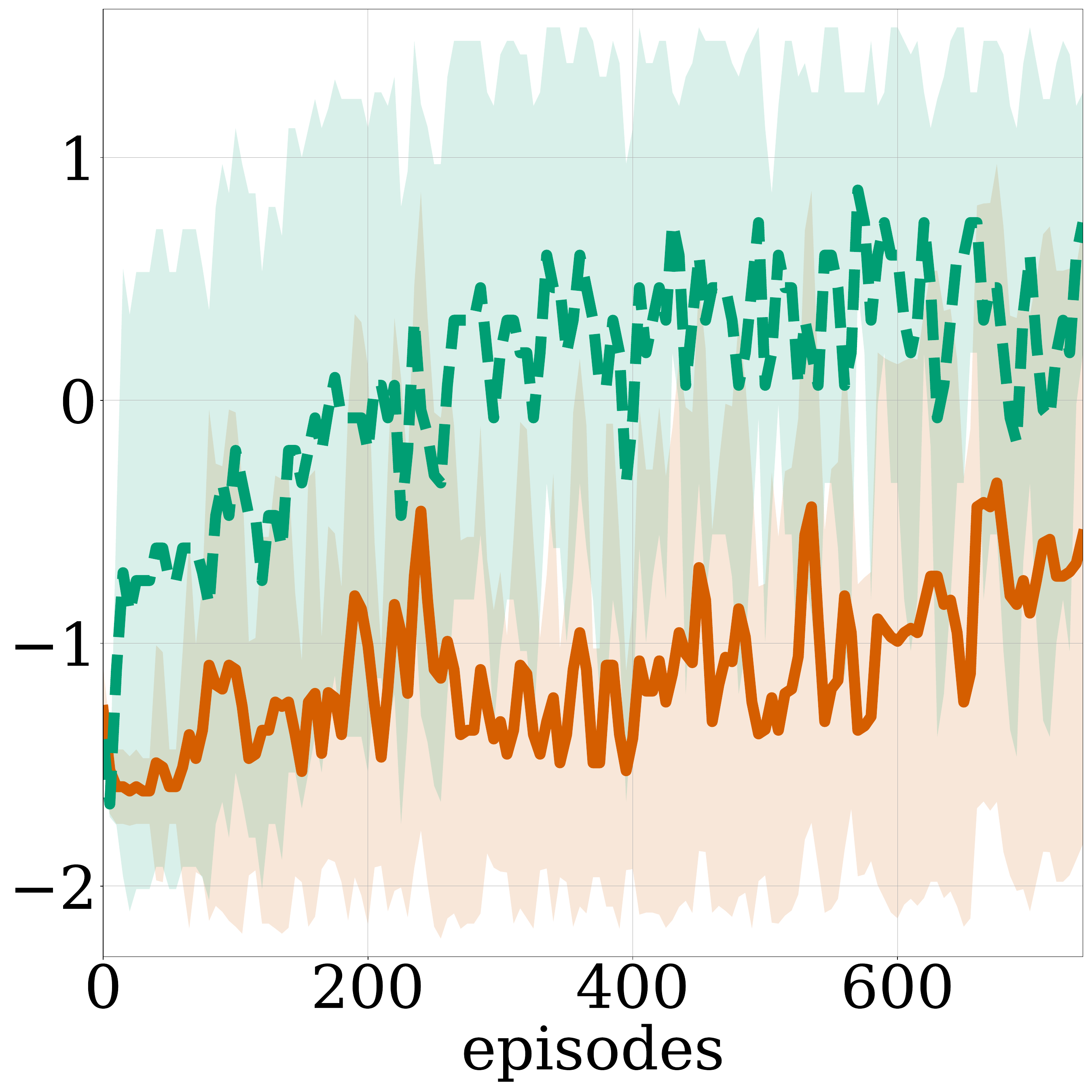}}\\
    \setcounter{subfigure}{0}
    \subfigure[Stack-2]{\includegraphics[width=0.19\textwidth]{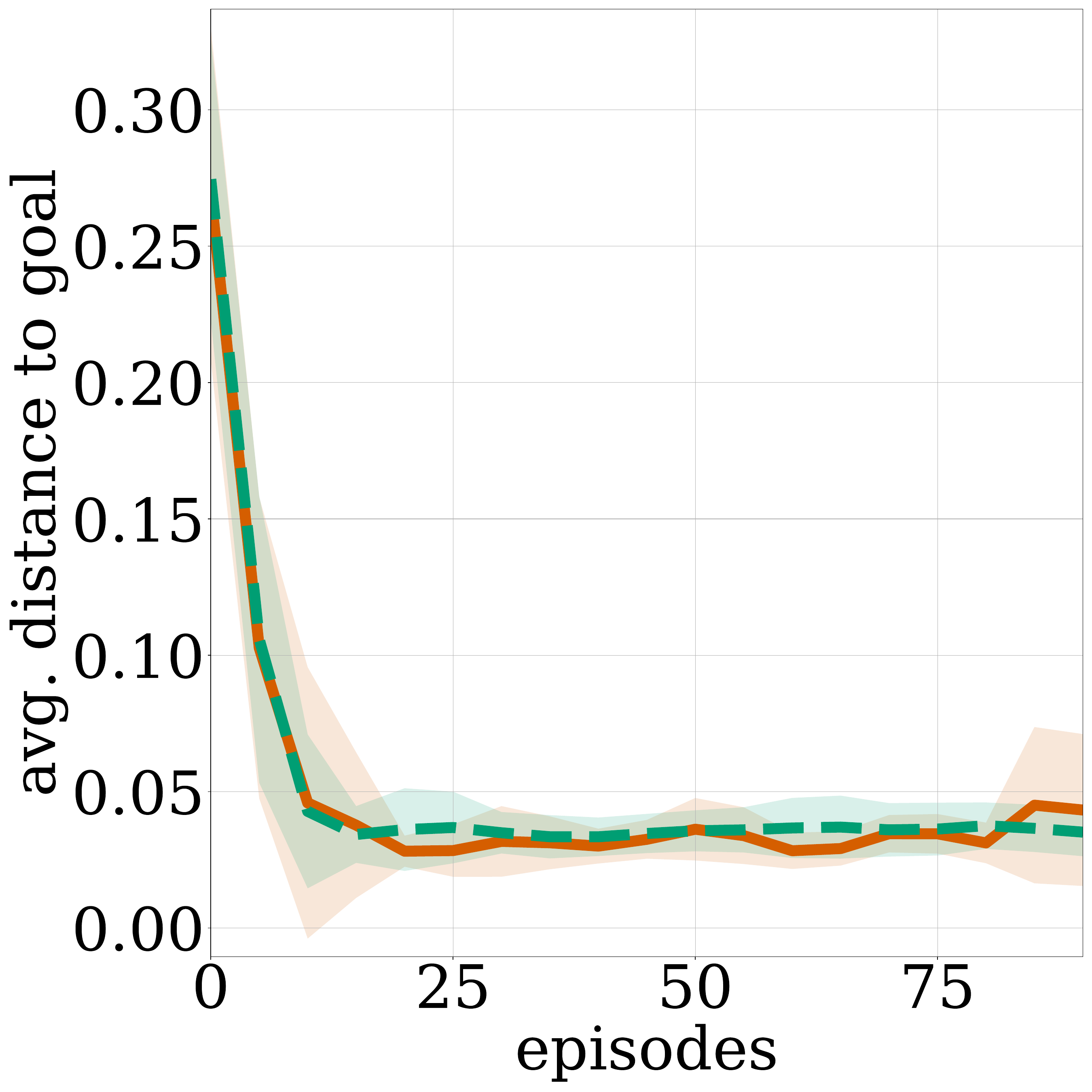}}
    \subfigure[Place-2]{\includegraphics[width=0.19\textwidth]{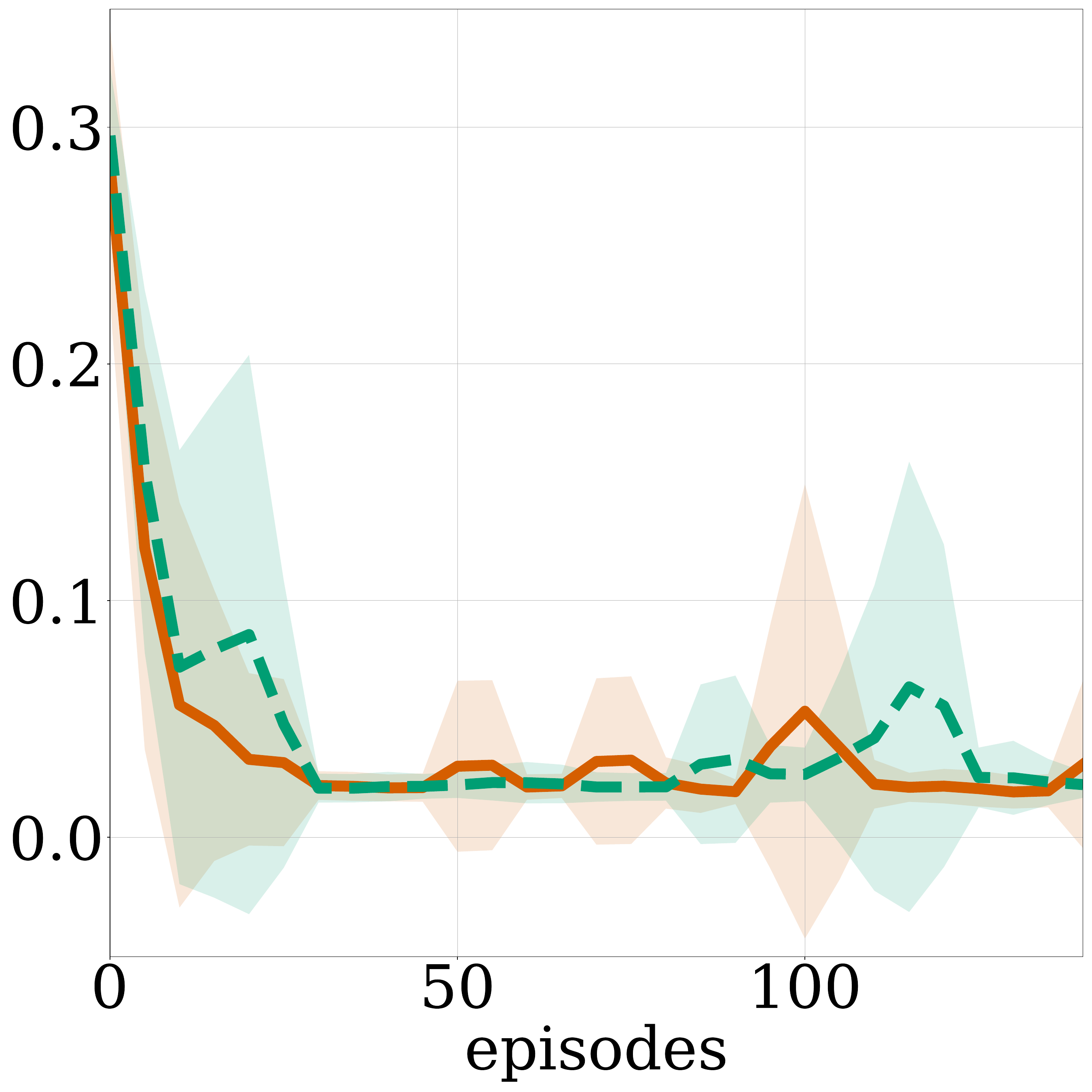}}
    \subfigure[Pyramid-3]{\includegraphics[width=0.19\textwidth]{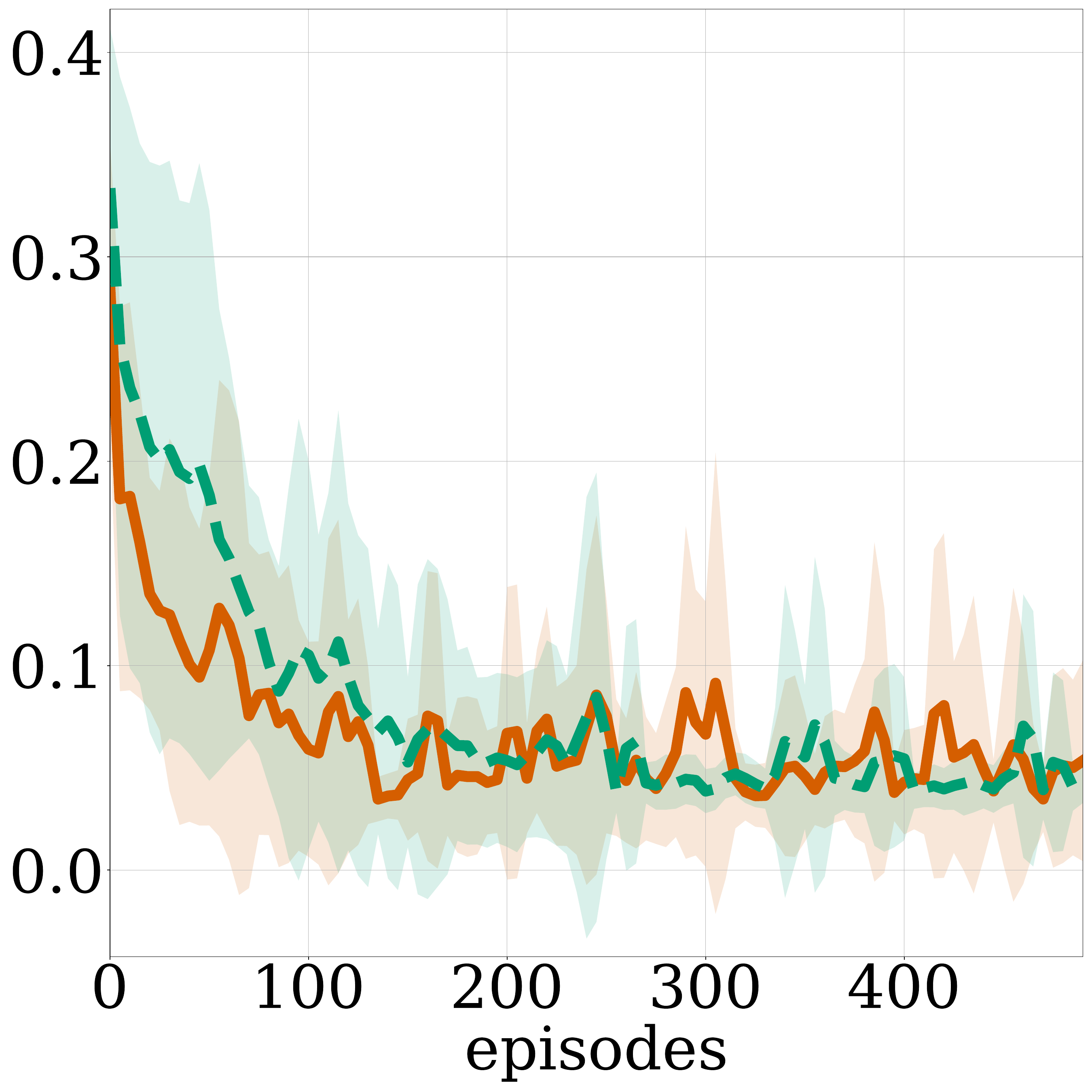}}
    \subfigure[Place-3]{\includegraphics[width=0.19\textwidth]{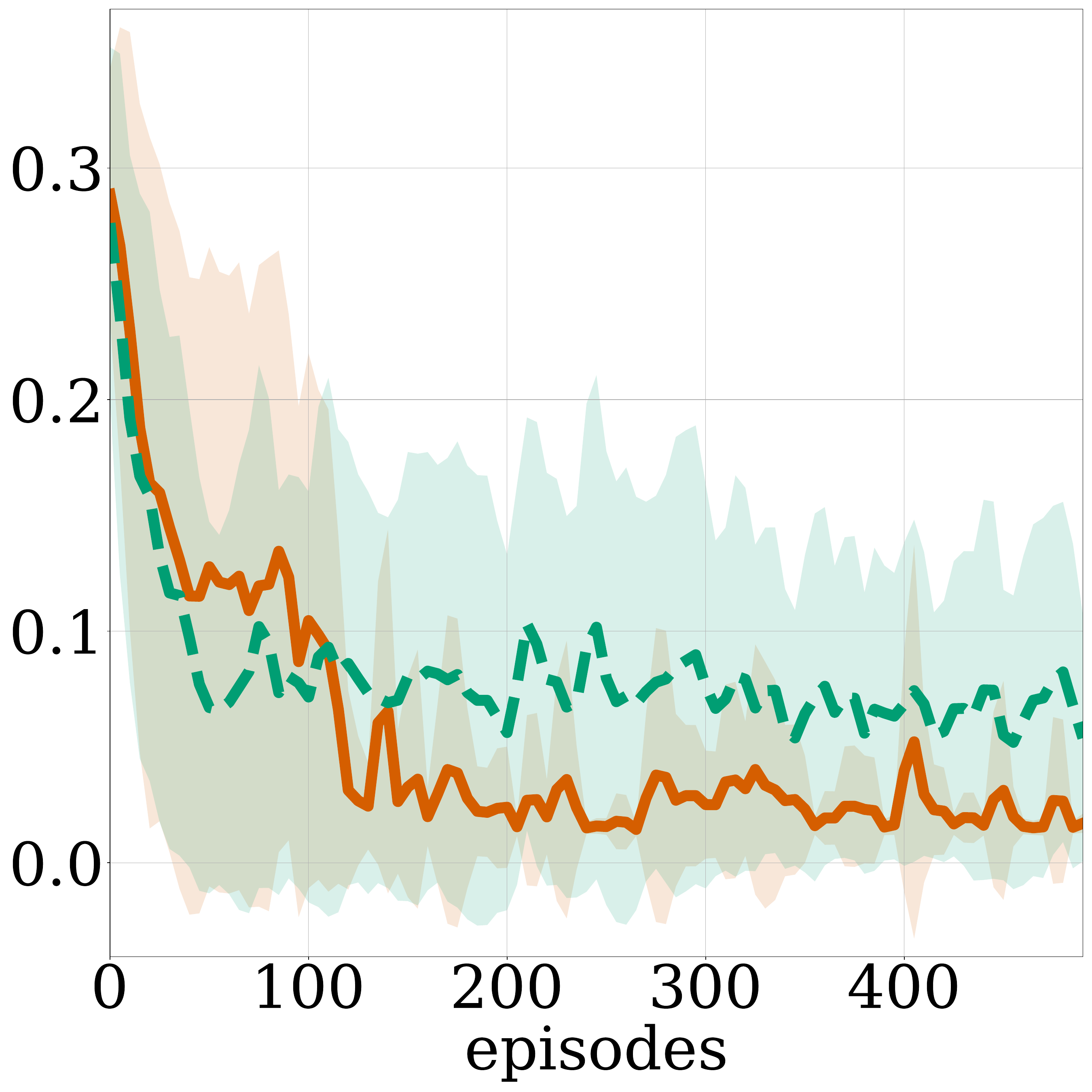}}
    \subfigure[Stack-3]{\includegraphics[width=0.19\textwidth]{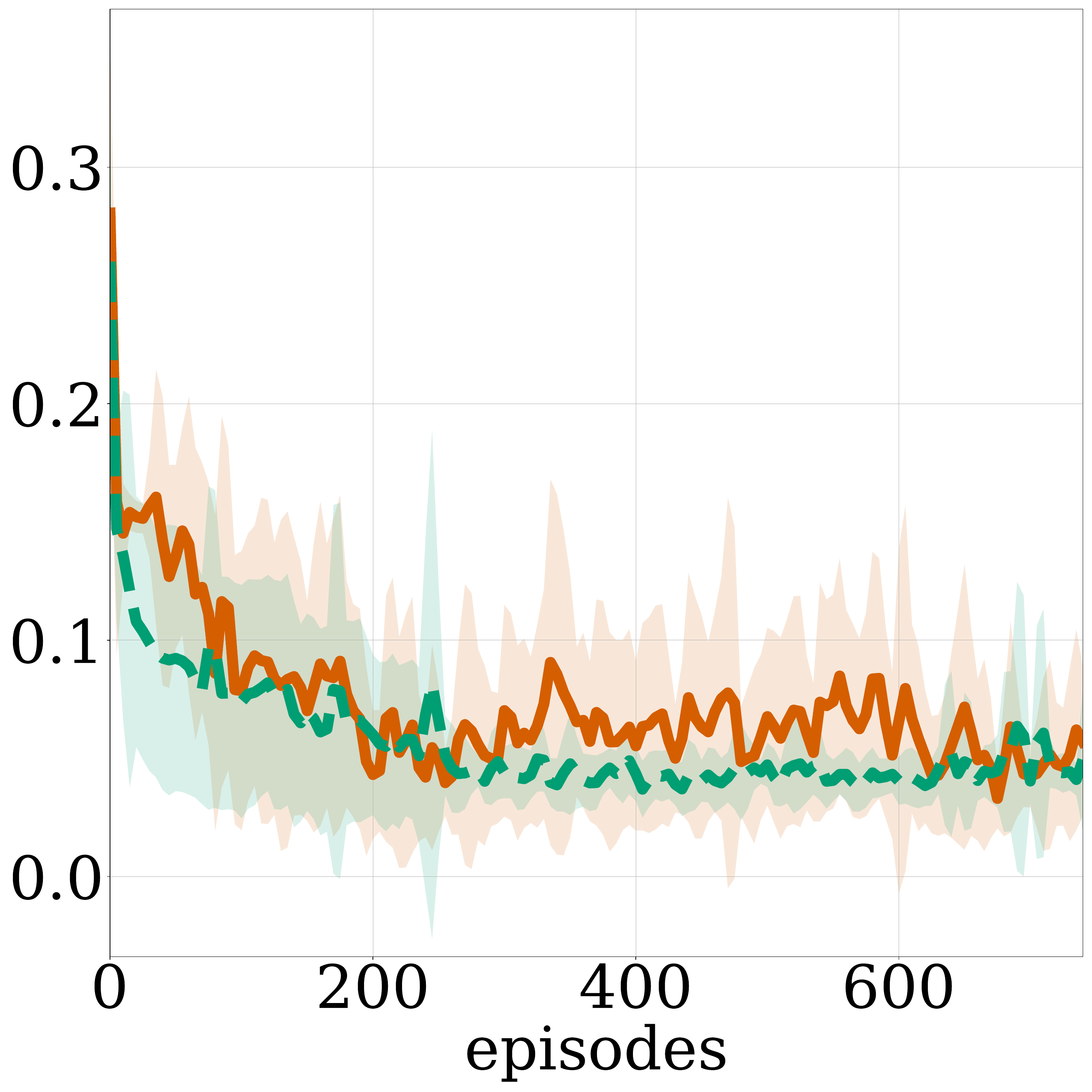}}
    \subfigure{\includegraphics[width=0.6\textwidth]{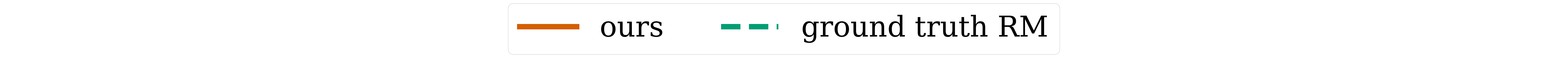}}\\
    \caption{Total reward obtained during one episode (top) and average distance between each object and its ground goal location, i.e. placement error (bottom) during training. Results are averaged over 10 runs, the shaded region represents the standard deviation.}
    \label{fig:quant_results}
\end{center}
\end{figure*}
\ \\
\\
Our method successfully inferred the correct RMs across all tasks, producing meaningful prototype states representing the demonstrated task. The inferred RM accommodates variations in sub-goal sequences observed in the demonstrations by representing each possible ordering as an alternative route through the RM states. Fig. \ref{fig:3b_2} shows the inferred RM and prototypes for the Place-3 task. As presented in the previous sections, each RM state corresponds to a unique high-level proposition that triggers a transition. The $(0)$-prototype represents the initial RM state, with each subsequent prototype corresponding to a specific task sub-goal. Fig. \ref{fig:rm-inference-overview} and Fig. \ref{fig:3b_3} depict the inferred RM and prototypes for the Pyramid-3 and the Stack-3 scenarios respectively.
\\
\\
To quantitatively assess our approach, we report both total reward per episode and placement error, measured as the average distance between each object and its goal position (Fig. \ref{fig:quant_results}). The total reward indicates if an agent effectively learns an optimal policy for the inferred RM. The placement error assesses the accuracy of the inferred RM, particularly in terms of its defined sub-goals. We compared our method to an agent using a ground truth RM with predefined propositions. The set of propositions is again defined such that for each state, one proposition becomes true. Each proposition is defined by the ground truth object positions corresponding to that sub-goal. A proposition becomes true if the placement error between the proposition's object positions and the current object positions is below 0.05m. It is important to note that the ground truth reward machine can only be utilized if the location of each object can be determined at every time step. Fortunately, this condition is met in our simulation environment. Every five episodes, we sampled a trajectory from the current greedy policy ($\epsilon = 0$) to measure total reward and placement error.
\\
\\
In achieving the maximum total reward of 1 per episode, agents must consistently transition towards and reach the final goal state in the RM graph (see Eq. \ref{eq:rm_reward}). Tasks with two blocks generally reached higher total reward levels compared to three-block tasks. Both agents, using either the inferred or ground truth RM, exhibit similar learning curves across most environments, except for Stack-3. In Stack-3, the agent trained on the ground truth RM achieved a higher reward plateau. Placement error trends were also similar between the two agents; however, for Place-3, the agent with the inferred RM achieved a lower placement error. All learning curves indicate that both agents’ DQNs converge to an optimal policy. Placement error comparisons show parity between our inferred RM and the ground truth RM agent, suggesting that any remaining error is primarily due to the control algorithm handling the pick-and-place actions. Both methods show notable fluctuations in reward and placement error, attributed to noise in executing robot primitives. In our approach, additional variance is introduced by prototype matching, where errors in feature embeddings occasionally prevent RM transitions despite correct block placements. This can cause agents to stall in a particular RM state until the episode ends. Despite these occasional instabilities, the low placement error demonstrates that an optimal policy is learned. It is noteworthy that for the Place-3 task, our method reaches a lower placement error. This is attributed to the fact that two runs for the ground truth RM agent, failed to learn an optimal policy.

\section{Conclusion}
We presented a novel method for inferring RMs from unstructured demonstrations, using video recordings as input. Unlike existing approaches, our method does not rely on predefined propositions. Instead, it uses clustering to derive meaningful sub-goals represented by prototypical states. By measuring the distance in feature space between the current observation and each prototype, our method detects sub-goal completion and enables state transitions within the RM. Experimental results show that our approach accurately infers the ground-truth RM with interpretable prototypes. Furthermore, the inferred RM enables an RL agent to learn an optimal policy, achieving similar placement accuracy to agents with access to ground-truth object positions and RMs.
\\
Currently, our method converges on a single policy path between the initial and goal states, even when multiple valid paths exist in the RM. A promising direction for future work is to develop agents that can adaptively select alternative paths, which would improve robustness in environments where certain paths become infeasible due to perturbations (e.g., a block becomes unavailable). An interesting body of work in this direction is maximum entropy RL \cite{haarnoja2018soft}. Additionally, enhancing prototype quality and detection accuracy could be achieved by exploring alternative embedding types, such as object-centric representations \cite{locatello2020object} that are likely more suited to robotic manipulation tasks. This could also facilitate the use of more complex types of objects. Integrating multi-camera views, or projecting views (e.g., from front to top), could further enrich feature representations, especially for real-world scenarios where camera placement may be limited. Finally, exploring the application of this technique in other domains, such as navigation, could extend its impact beyond manipulation tasks.


\section{Acknowledgement}
This research was partially funded by the Flemish Government (Flanders AI Research Program).

\bibliography{aaai24}

\begin{thebibliography}{41}
\providecommand{\natexlab}[1]{#1}

\bibitem[{Abbeel, Coates, and Ng(2010)}]{abbeel2010autonomous}
Abbeel, P.; Coates, A.; and Ng, A.~Y. 2010.
\newblock Autonomous helicopter aerobatics through apprenticeship learning.
\newblock \emph{The International Journal of Robotics Research}, 29(13): 1608--1639.

\bibitem[{Abbeel and Ng(2004)}]{abbeel2004apprenticeship}
Abbeel, P.; and Ng, A.~Y. 2004.
\newblock Apprenticeship learning via inverse reinforcement learning.
\newblock In \emph{Proceedings of the twenty-first international conference on Machine learning}, 1.

\bibitem[{Araki et~al.(2021)Araki, Li, Vodrahalli, DeCastro, Fry, and Rus}]{araki2021logical}
Araki, B.; Li, X.; Vodrahalli, K.; DeCastro, J.; Fry, M.; and Rus, D. 2021.
\newblock The logical options framework.
\newblock In \emph{International Conference on Machine Learning}, 307--317. PMLR.

\bibitem[{Araki et~al.(2019)Araki, Vodrahalli, Leech, Vasile, Donahue, and Rus}]{araki2019learning}
Araki, B.; Vodrahalli, K.; Leech, T.; Vasile, C.-I.; Donahue, M.~D.; and Rus, D.~L. 2019.
\newblock Learning to plan with logical automata.

\bibitem[{Argall et~al.(2009)Argall, Chernova, Veloso, and Browning}]{argall2009survey}
Argall, B.~D.; Chernova, S.; Veloso, M.; and Browning, B. 2009.
\newblock A survey of robot learning from demonstration.
\newblock \emph{Robotics and autonomous systems}, 57(5): 469--483.

\bibitem[{Baert, Leroux, and Simoens(2024{\natexlab{a}})}]{baert2024learningtask}
Baert, M.; Leroux, S.; and Simoens, P. 2024{\natexlab{a}}.
\newblock Learning Task Specifications from Demonstrations as Probabilistic Automata.
\newblock \emph{arXiv preprint arXiv:2409.07091}.

\bibitem[{Baert, Leroux, and Simoens(2024{\natexlab{b}})}]{baert2024learning}
Baert, M.; Leroux, S.; and Simoens, P. 2024{\natexlab{b}}.
\newblock Learning Temporal Task Specifications From Demonstrations.
\newblock In \emph{International Workshop on Explainable, Transparent Autonomous Agents and Multi-Agent Systems}, 81--98. Springer.

\bibitem[{Baert et~al.(2023)Baert, Mazzaglia, Leroux, and Simoens}]{baert2023maximum}
Baert, M.; Mazzaglia, P.; Leroux, S.; and Simoens, P. 2023.
\newblock Maximum Causal Entropy Inverse Constrained Reinforcement Learning.
\newblock \emph{arXiv preprint arXiv:2305.02857}.

\bibitem[{Bellman(1957)}]{bellman1957markovian}
Bellman, R. 1957.
\newblock A Markovian decision process.
\newblock \emph{Journal of mathematics and mechanics}, 679--684.

\bibitem[{Bombara and Belta(2021)}]{bombara2021offline}
Bombara, G.; and Belta, C. 2021.
\newblock Offline and online learning of signal temporal logic formulae using decision trees.
\newblock \emph{ACM Transactions on Cyber-Physical Systems}, 5(3): 1--23.

\bibitem[{Camacho et~al.(2021)Camacho, Varley, Zeng, Jain, Iscen, and Kalashnikov}]{camacho2021reward}
Camacho, A.; Varley, J.; Zeng, A.; Jain, D.; Iscen, A.; and Kalashnikov, D. 2021.
\newblock Reward machines for vision-based robotic manipulation.
\newblock In \emph{2021 IEEE International Conference on Robotics and Automation (ICRA)}, 14284--14290. IEEE.

\bibitem[{Deng et~al.(2009)Deng, Dong, Socher, Li, Li, and Fei-Fei}]{deng2009imagenet}
Deng, J.; Dong, W.; Socher, R.; Li, L.-J.; Li, K.; and Fei-Fei, L. 2009.
\newblock Imagenet: A large-scale hierarchical image database.
\newblock In \emph{2009 IEEE conference on computer vision and pattern recognition}, 248--255. Ieee.

\bibitem[{Dohmen et~al.(2022)Dohmen, Topper, Atia, Beckus, Trivedi, and Velasquez}]{dohmen2022inferring}
Dohmen, T.; Topper, N.; Atia, G.; Beckus, A.; Trivedi, A.; and Velasquez, A. 2022.
\newblock Inferring probabilistic reward machines from non-markovian reward signals for reinforcement learning.
\newblock In \emph{Proceedings of the International Conference on Automated Planning and Scheduling}, volume~32, 574--582.

\bibitem[{Ester et~al.(1996)Ester, Kriegel, Sander, Xu et~al.}]{ester1996density}
Ester, M.; Kriegel, H.-P.; Sander, J.; Xu, X.; et~al. 1996.
\newblock A density-based algorithm for discovering clusters in large spatial databases with noise.
\newblock In \emph{kdd}, volume~96, 226--231.

\bibitem[{Ghazanfari, Afghah, and Taylor(2020)}]{ghazanfari2020sequential}
Ghazanfari, B.; Afghah, F.; and Taylor, M.~E. 2020.
\newblock Sequential association rule mining for autonomously extracting hierarchical task structures in reinforcement learning.
\newblock \emph{IEEE Access}, 8: 11782--11799.

\bibitem[{Haarnoja et~al.(2018)Haarnoja, Zhou, Abbeel, and Levine}]{haarnoja2018soft}
Haarnoja, T.; Zhou, A.; Abbeel, P.; and Levine, S. 2018.
\newblock Soft actor-critic: Off-policy maximum entropy deep reinforcement learning with a stochastic actor.
\newblock In \emph{International conference on machine learning}, 1861--1870. PMLR.

\bibitem[{He et~al.(2015)He, Zhang, Ren, and Sun}]{he2015deep}
He, K.; Zhang, X.; Ren, S.; and Sun, J. 2015.
\newblock Deep residual learning for image recognition. CoRR abs/1512.03385 (2015).

\bibitem[{Hester et~al.(2018)Hester, Vecerik, Pietquin, Lanctot, Schaul, Piot, Horgan, Quan, Sendonaris, Osband et~al.}]{hester2018deep}
Hester, T.; Vecerik, M.; Pietquin, O.; Lanctot, M.; Schaul, T.; Piot, B.; Horgan, D.; Quan, J.; Sendonaris, A.; Osband, I.; et~al. 2018.
\newblock Deep q-learning from demonstrations.
\newblock In \emph{Proceedings of the AAAI conference on artificial intelligence}, volume~32.

\bibitem[{Ho and Ermon(2016)}]{ho2016generative}
Ho, J.; and Ermon, S. 2016.
\newblock Generative adversarial imitation learning.
\newblock \emph{Advances in neural information processing systems}, 29.

\bibitem[{Icarte et~al.(2022)Icarte, Klassen, Valenzano, and McIlraith}]{icarte2022reward}
Icarte, R.~T.; Klassen, T.~Q.; Valenzano, R.; and McIlraith, S.~A. 2022.
\newblock Reward machines: Exploiting reward function structure in reinforcement learning.
\newblock \emph{Journal of Artificial Intelligence Research}, 73: 173--208.

\bibitem[{Kingma(2014)}]{kingma2014adam}
Kingma, D.~P. 2014.
\newblock Adam: A method for stochastic optimization.
\newblock \emph{arXiv preprint arXiv:1412.6980}.

\bibitem[{Kong, Jones, and Belta(2016)}]{kong2016temporal}
Kong, Z.; Jones, A.; and Belta, C. 2016.
\newblock Temporal logics for learning and detection of anomalous behavior.
\newblock \emph{IEEE Transactions on Automatic Control}, 62(3): 1210--1222.

\bibitem[{Kuo, Katz, and Barbu(2020)}]{kuo2020encoding}
Kuo, Y.-L.; Katz, B.; and Barbu, A. 2020.
\newblock Encoding formulas as deep networks: Reinforcement learning for zero-shot execution of LTL formulas.
\newblock In \emph{2020 IEEE/RSJ International Conference on Intelligent Robots and Systems (IROS)}, 5604--5610. IEEE.

\bibitem[{Li, Vasile, and Belta(2017)}]{li2017reinforcement}
Li, X.; Vasile, C.-I.; and Belta, C. 2017.
\newblock Reinforcement learning with temporal logic rewards.
\newblock In \emph{2017 IEEE/RSJ International Conference on Intelligent Robots and Systems (IROS)}, 3834--3839. IEEE.

\bibitem[{Locatello et~al.(2020)Locatello, Weissenborn, Unterthiner, Mahendran, Heigold, Uszkoreit, Dosovitskiy, and Kipf}]{locatello2020object}
Locatello, F.; Weissenborn, D.; Unterthiner, T.; Mahendran, A.; Heigold, G.; Uszkoreit, J.; Dosovitskiy, A.; and Kipf, T. 2020.
\newblock Object-centric learning with slot attention.
\newblock \emph{Advances in neural information processing systems}, 33: 11525--11538.

\bibitem[{Long, Shelhamer, and Darrell(2015)}]{long2015fully}
Long, J.; Shelhamer, E.; and Darrell, T. 2015.
\newblock Fully convolutional networks for semantic segmentation.
\newblock In \emph{Proceedings of the IEEE conference on computer vision and pattern recognition}, 3431--3440.

\bibitem[{Mandlekar et~al.(2020)Mandlekar, Xu, Mart{\'{\i}}n{-}Mart{\'{\i}}n, Savarese, and Fei{-}Fei}]{DBLP:conf/rss/MandlekarXMS020}
Mandlekar, A.; Xu, D.; Mart{\'{\i}}n{-}Mart{\'{\i}}n, R.; Savarese, S.; and Fei{-}Fei, L. 2020.
\newblock {GTI:} Learning to Generalize across Long-Horizon Tasks from Human Demonstrations.
\newblock In Toussaint, M.; Bicchi, A.; and Hermans, T., eds., \emph{Robotics: Science and Systems XVI, Virtual Event / Corvalis, Oregon, USA, July 12-16, 2020}.

\bibitem[{Mnih et~al.(2015)Mnih, Kavukcuoglu, Silver, Rusu, Veness, Bellemare, Graves, Riedmiller, Fidjeland, Ostrovski et~al.}]{mnih2015human}
Mnih, V.; Kavukcuoglu, K.; Silver, D.; Rusu, A.~A.; Veness, J.; Bellemare, M.~G.; Graves, A.; Riedmiller, M.; Fidjeland, A.~K.; Ostrovski, G.; et~al. 2015.
\newblock Human-level control through deep reinforcement learning.
\newblock \emph{nature}, 518(7540): 529--533.

\bibitem[{Ng, Harada, and Russell(1999)}]{ng1999policy}
Ng, A.~Y.; Harada, D.; and Russell, S. 1999.
\newblock Policy invariance under reward transformations: Theory and application to reward shaping.
\newblock In \emph{Icml}, volume~99, 278--287.

\bibitem[{Roy et~al.(2023)Roy, Gaglione, Baharisangari, Neider, Xu, and Topcu}]{roy2023learning}
Roy, R.; Gaglione, J.-R.; Baharisangari, N.; Neider, D.; Xu, Z.; and Topcu, U. 2023.
\newblock Learning interpretable temporal properties from positive examples only.
\newblock In \emph{Proceedings of the AAAI Conference on Artificial Intelligence}, volume~37, 6507--6515.

\bibitem[{Shah et~al.(2018)Shah, Kamath, Shah, and Li}]{shah2018bayesian}
Shah, A.; Kamath, P.; Shah, J.~A.; and Li, S. 2018.
\newblock Bayesian inference of temporal task specifications from demonstrations.
\newblock \emph{Advances in Neural Information Processing Systems}, 31.

\bibitem[{Silver et~al.(2016)Silver, Huang, Maddison, Guez, Sifre, Van Den~Driessche, Schrittwieser, Antonoglou, Panneershelvam, Lanctot et~al.}]{silver2016mastering}
Silver, D.; Huang, A.; Maddison, C.~J.; Guez, A.; Sifre, L.; Van Den~Driessche, G.; Schrittwieser, J.; Antonoglou, I.; Panneershelvam, V.; Lanctot, M.; et~al. 2016.
\newblock Mastering the game of Go with deep neural networks and tree search.
\newblock \emph{nature}, 529(7587): 484--489.

\bibitem[{Toro~Icarte et~al.(2019)Toro~Icarte, Waldie, Klassen, Valenzano, Castro, and McIlraith}]{toro2019learning}
Toro~Icarte, R.; Waldie, E.; Klassen, T.; Valenzano, R.; Castro, M.; and McIlraith, S. 2019.
\newblock Learning reward machines for partially observable reinforcement learning.
\newblock \emph{Advances in neural information processing systems}, 32.

\bibitem[{Verginis et~al.(2024)Verginis, Koprulu, Chinchali, and Topcu}]{verginis2024joint}
Verginis, C.~K.; Koprulu, C.; Chinchali, S.; and Topcu, U. 2024.
\newblock Joint learning of reward machines and policies in environments with partially known semantics.
\newblock \emph{Artificial Intelligence}, 104146.

\bibitem[{Voloshin et~al.(2022)Voloshin, Le, Chaudhuri, and Yue}]{voloshin2022policy}
Voloshin, C.; Le, H.; Chaudhuri, S.; and Yue, Y. 2022.
\newblock Policy optimization with linear temporal logic constraints.
\newblock \emph{Advances in Neural Information Processing Systems}, 35: 17690--17702.

\bibitem[{Xiong et~al.(2022)Xiong, Eappen, Qureshi, and Jagannathan}]{xiong2022constrained}
Xiong, Z.; Eappen, J.; Qureshi, A.~H.; and Jagannathan, S. 2022.
\newblock Constrained Hierarchical Deep Reinforcement Learning with Differentiable Formal Specifications.

\bibitem[{Xu et~al.(2020)Xu, Gavran, Ahmad, Majumdar, Neider, Topcu, and Wu}]{xu2020joint}
Xu, Z.; Gavran, I.; Ahmad, Y.; Majumdar, R.; Neider, D.; Topcu, U.; and Wu, B. 2020.
\newblock Joint inference of reward machines and policies for reinforcement learning.
\newblock In \emph{Proceedings of the International Conference on Automated Planning and Scheduling}, volume~30, 590--598.

\bibitem[{Xu et~al.(2021)Xu, Wu, Ojha, Neider, and Topcu}]{xu2021active}
Xu, Z.; Wu, B.; Ojha, A.; Neider, D.; and Topcu, U. 2021.
\newblock Active finite reward automaton inference and reinforcement learning using queries and counterexamples.
\newblock In \emph{Machine Learning and Knowledge Extraction: 5th IFIP TC 5, TC 12, WG 8.4, WG 8.9, WG 12.9 International Cross-Domain Conference, CD-MAKE 2021, Virtual Event, August 17--20, 2021, Proceedings 5}, 115--135. Springer.

\bibitem[{Zeng et~al.(2022)Zeng, Song, Yu, Donlon, Hogan, Bauza, Ma, Taylor, Liu, Romo et~al.}]{zeng2022robotic}
Zeng, A.; Song, S.; Yu, K.-T.; Donlon, E.; Hogan, F.~R.; Bauza, M.; Ma, D.; Taylor, O.; Liu, M.; Romo, E.; et~al. 2022.
\newblock Robotic pick-and-place of novel objects in clutter with multi-affordance grasping and cross-domain image matching.
\newblock \emph{The International Journal of Robotics Research}, 41(7): 690--705.

\bibitem[{Zhang et~al.(2018)Zhang, McCarthy, Jow, Lee, Chen, Goldberg, and Abbeel}]{DBLP:conf/icra/ZhangMJLCGA18}
Zhang, T.; McCarthy, Z.; Jow, O.; Lee, D.; Chen, X.; Goldberg, K.; and Abbeel, P. 2018.
\newblock Deep Imitation Learning for Complex Manipulation Tasks from Virtual Reality Teleoperation.
\newblock In \emph{2018 {IEEE} International Conference on Robotics and Automation, {ICRA} 2018, Brisbane, Australia, May 21-25, 2018}, 1--8. {IEEE}.

\bibitem[{Zhu et~al.(2020)Zhu, Wong, Mandlekar, Mart\'{i}n-Mart\'{i}n, Joshi, Nasiriany, and Zhu}]{robosuite2020}
Zhu, Y.; Wong, J.; Mandlekar, A.; Mart\'{i}n-Mart\'{i}n, R.; Joshi, A.; Nasiriany, S.; and Zhu, Y. 2020.
\newblock robosuite: A Modular Simulation Framework and Benchmark for Robot Learning.
\newblock In \emph{arXiv preprint arXiv:2009.12293}.

\end{thebibliography}

\end{document}